# Retrieval and Clustering from a 3D Human Database based on Body and Head Shape

**Afzal Godil, Sandy Ressler**
National Institute of Standards and Technology

## ABSTRACT

In this paper, we describe a framework for similarity based retrieval and clustering from a 3D human database. Our technique is based on both body and head shape representation and the retrieval is based on similarity of both of them. The 3D human database used in our study is the CAESAR anthropometric database which contains approximately 5000 bodies. We have developed a web-based interface for specifying the queries to interact with the retrieval system. Our approach performs the similarity based retrieval in a reasonable amount of time and is a practical approach.

## INTRODUCTION

With the wide availability of 3D scanning technologies, 3D geometry is becoming an important type of media. Large numbers of 3D models are created every day and many are stored in publicly available databases. Understanding the 3D shape and structure of these models is essential to many scientific activities. These 3D scientific databases require methods for storage, indexing, searching, clustering, retrieval and recognition of the content under study. Searching a database for 3D objects which are similar to a given 3D object is an important and challenging problem. This domain of research is also called query by example (QBE). We have developed techniques for searching a human database and have used the CAESAR anthropometric database which consists of a database of approximately 5000 human subjects. In our study we have implemented methods for similarity based retrieval from the CAESAR human database based on both human body and head shape.

Previous work on human body retrieval based on body shape was performed by [Paquet and Rioux 2003]. They performed content-based anthropometric data mining of three dimensional scanned human by representing them with compact support feature vectors. They showed a virtual environment to perform visual data mining on the clusters and to characterize the population by defining archetypes.

[Paquet 2004] introduced cluster analysis as a method to explore 3D body scans together with the relational anthropometric data as contained in the CAESAR anthropometric database. [Azouz 2002, 2004] analyzed human shape variability using a volumetric representation of 3D human bodies and applied a principal components analysis (PCA) to the volumetric data to extract dominant components of shape variability for a target population. Through visualization, they also showed the main modes of human shape variation. The work of [Allen 2004] demonstrated a system of synthesizing 3D human body shapes, according to user specified parameters; they used 250 CAESAR body scans for training.

Retrieval based on head shape was performed by [Ip and Wong 2002]. Their similarity measure was based on Extended Gaussian Images of the polygon normal. They also compared it to an Eigenhead approach.

The 3D scans of human bodies in the CAESAR human database contain over two hundred fifty thousand grid points. To be used effectively for indexing, searching, clustering and retrieval, this human body data requires a compact representation. We have developed two such representations based on human body shape: 1) a descriptor vector d, based on lengths mostly between single large bones. Thus, we form a 3D body description vector of fifteen distances, d, with wrist to elbow, elbow to shoulder, hip to knee etc.; and 2) three silhouettes of the human body are created by rendering the human body from the front, side and top. These silhouettes are then encoded as Fourier descriptors of features for later similarity based retrieval. These two methods are explained in more details in the Body Shape Descriptor section.

We also have developed two compact representation based on human head shape: 1) applying Principal Component Analysis (PCA) on the 3D facial surface and creating PCA based facial descriptors; and 2) in the second method the 3D triangular grid of the head is transformed to a spherical coordinate system by a least squares approach and expanded in a basis of spherical harmonics. More explanation of these two representations of human head shape follow in the Head shape Descriptor section.

We also have used these four descriptors for clustering of human bodies based on each descriptor. The four descriptors allow the selection of the best descriptor for the application, such as the use of a head descriptor for Helmet Design.

## CAESAR database

The CAESAR (Civilian American and European Surface Anthropometry Resource) project has collected 3D Scans, seventy-three Anthropometry Landmarks, and Traditional Measurements data for each of the 5000 subjects. The objective of this study was to represent, in three-dimensions, the anthropometric variability of the civilian populations of Europe and North America and it was the first successful anthropometric survey to use 3-D scanning technology. The CAESAR project employs both 3-D scanning and traditional tools for body measurements for people ages 18-65. A typical CAESAR body is shown in Figure 1.

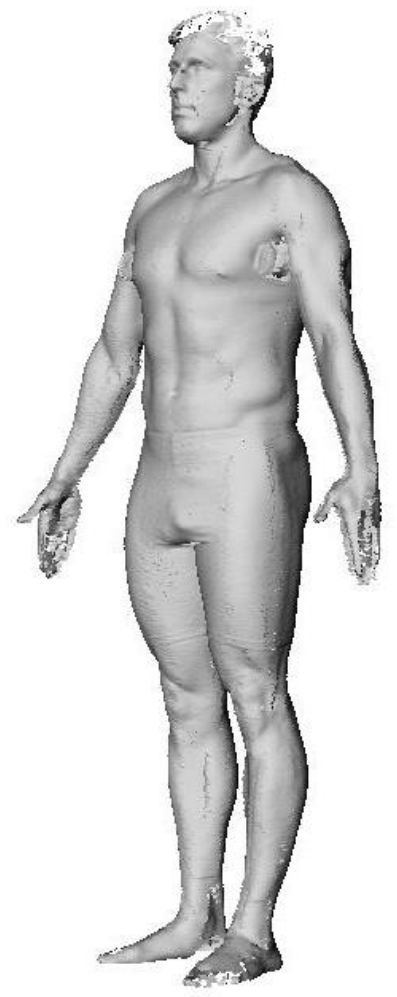

Figure 1a. A CAESAR body in standing pose

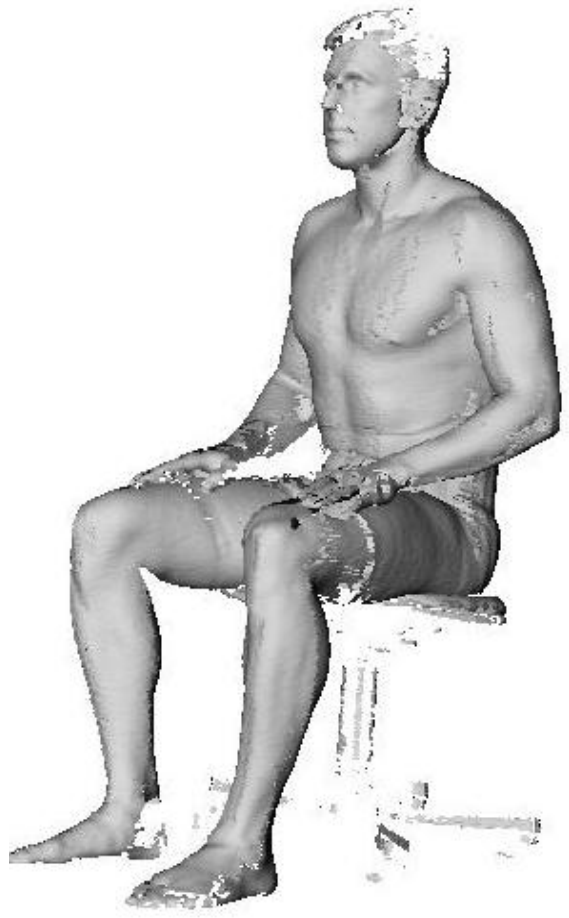

Figure 1b. A CAESAR body in sitting pose

The seventy-three anthropometric landmarks points were extracted from the scans as shown in Figure 2. These landmark points are pre-marked by pasting small stickers on the body and are automatically extracted using landmark software. There are around 250,000 points in each surface grid on a body and points are distributed uniformly.

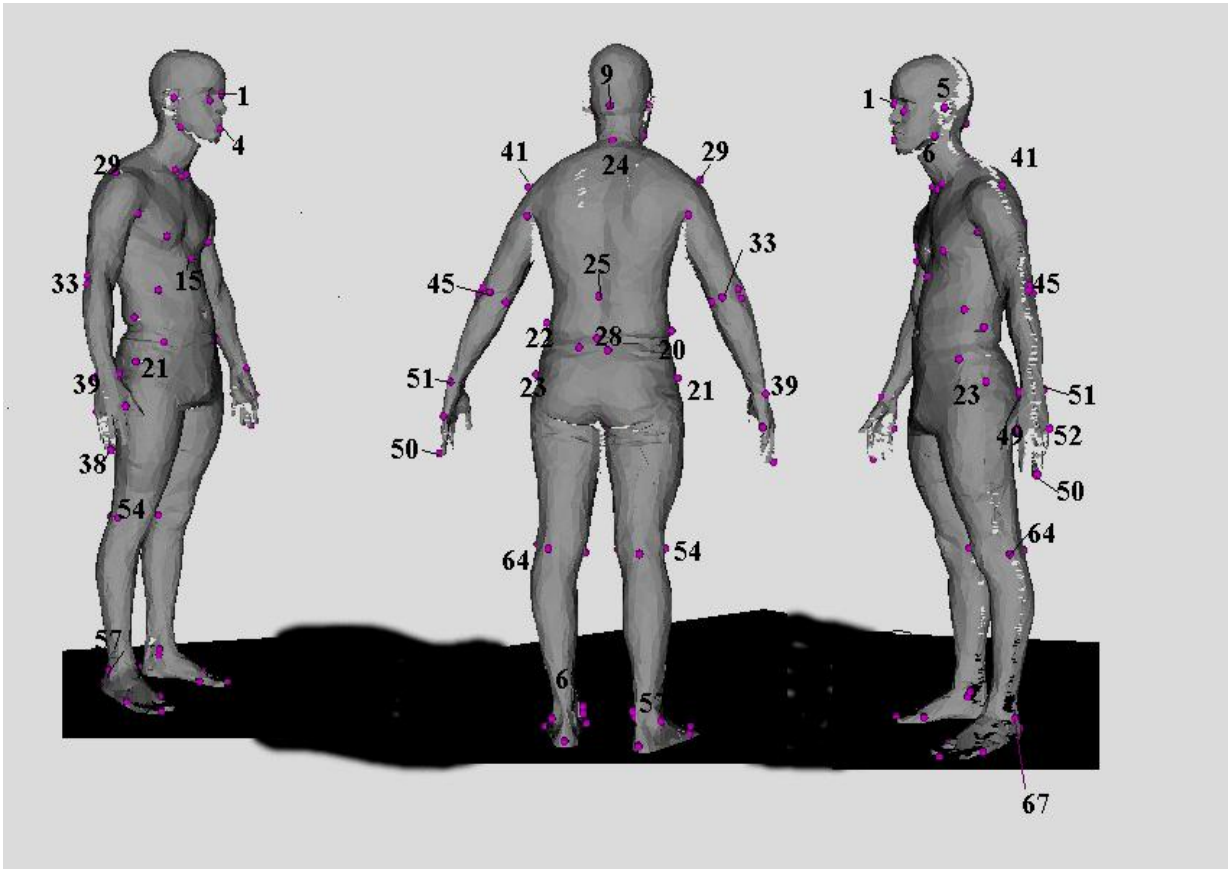

Figure 2. A Caesar body with landmark numbers and positions

## BODY SHAPE DESCRIPTOR

We now describe two methods for creating descriptors based on human body shape:

### Distance Based

The first method uses a descriptor vector d based on lengths mostly between single large bones. For descriptor vector purposes, we require lengths only between landmark points where their separation distance is somewhat pose-independent. The reason it is not completely pose invariant is that distance are between landmark points which are on the surface body compared to the distance between the center of the joint axis. This applies to points connected by a single large bone as shown in Figure 3. Thus, we form a descriptor vector of fifteen distances, d, with d1 wrist to elbow, d2, elbow to shoulder, d3 hip to knee etc. For which the Euclidean distance

$$\mathbf{d} = ||Pi - Pj||$$

is somewhat invariant across different poses. Distances such as chin-knee are avoided. The distance based descriptor is then used with the L1 and L2 norm to create a similarity matrix.

The L1 distance: $d\left(P_i, P_j\right) = \sum_{i=1}^{k} \left| P_i - P_j \right|$

The L2 distance: $d\left(P_i, P_j\right) = \sum_{i=1}^{k} \left| \sqrt{P_i^2 - P_j^2} \right|$

More details and shortcomings about this descriptor were described in the paper [Godil 2003]

To test how well the distance based descriptor performs, we studied the identification rate of a subset of 200

subjects of CESAR database where the gallery set contains the standing and the probe set contains the sitting pose of each subject. In this discussion, the gallery is the group of enrolled descriptor vector and the probe set refers to the group of unknown test descriptor vectors.

The measure of identification performance is the “rank order statistic” called the Cumulative Match Characteristic (CMC). The rank order statistics indicates the probability that the gallery subject will be among the top r matches to a probe image of the same subject. This probability depends upon both gallery size and rank. The CMC at rank 1 for the study is 40%.

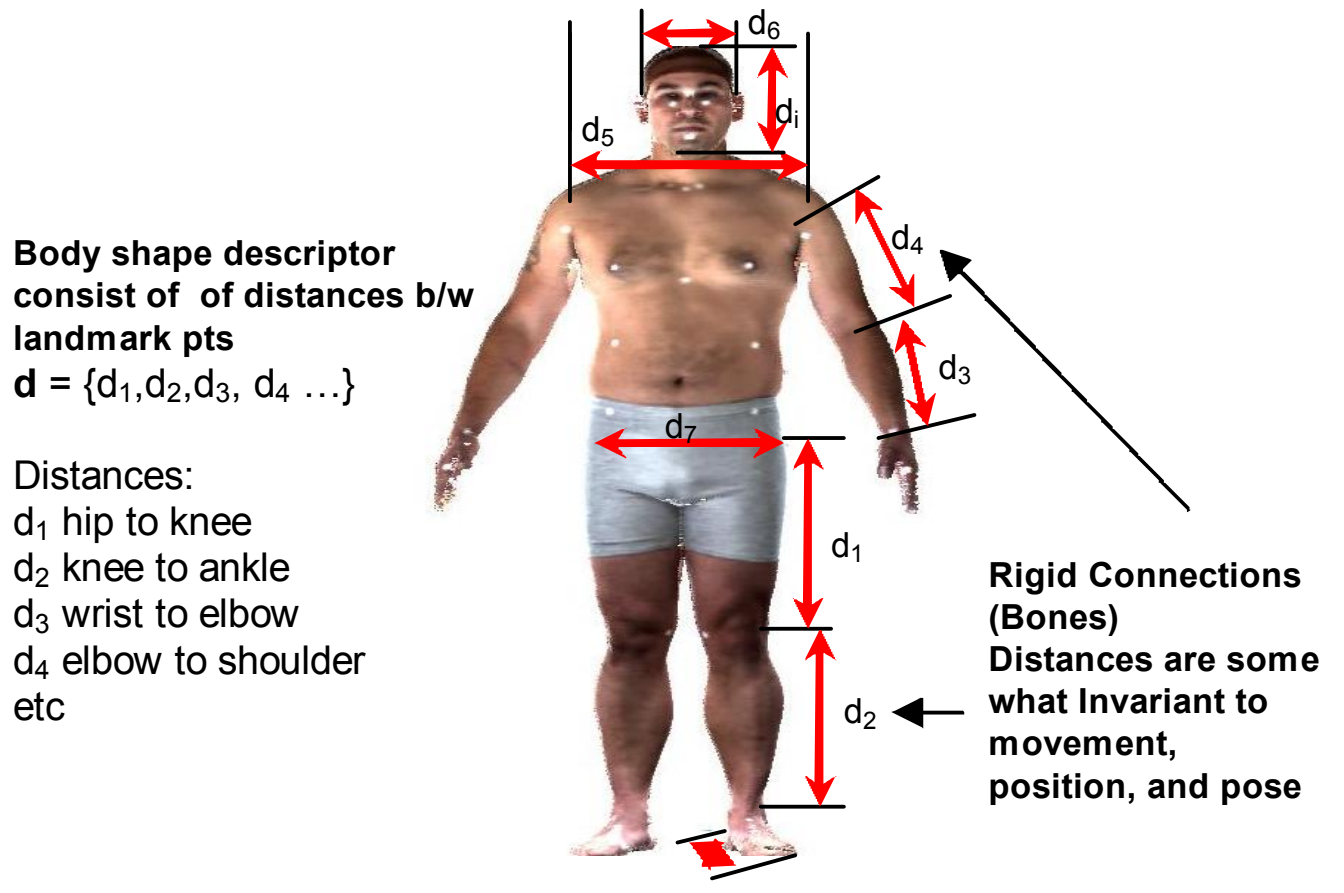

Figure 3. A distance based body shape descriptor

## Silhouette Fourier Based

The second method of body shape descriptor that we propose is based on rendering the human body from the front, side and top directions and creating three silhouettes of the human body as shown in Figure 4. The theory is that 3D models are similar if they also look similar from different viewing angles. The silhouette is then represented as R(radius) of the outer contour from the center of origin of the area of the silhouettes. These three contours are then encoded as Fourier descriptors which are used later as features for similarity based retrieval. The number of real part of Fourier modes used to describe each silhouette is sixteen (16); hence each human body is described by a vector of length forty eight (48). This method is pose dependent, so only bodies of the same pose can be compared. The Fourier based descriptor is then used with the L1 and L2 norm to create a similarity matrix.

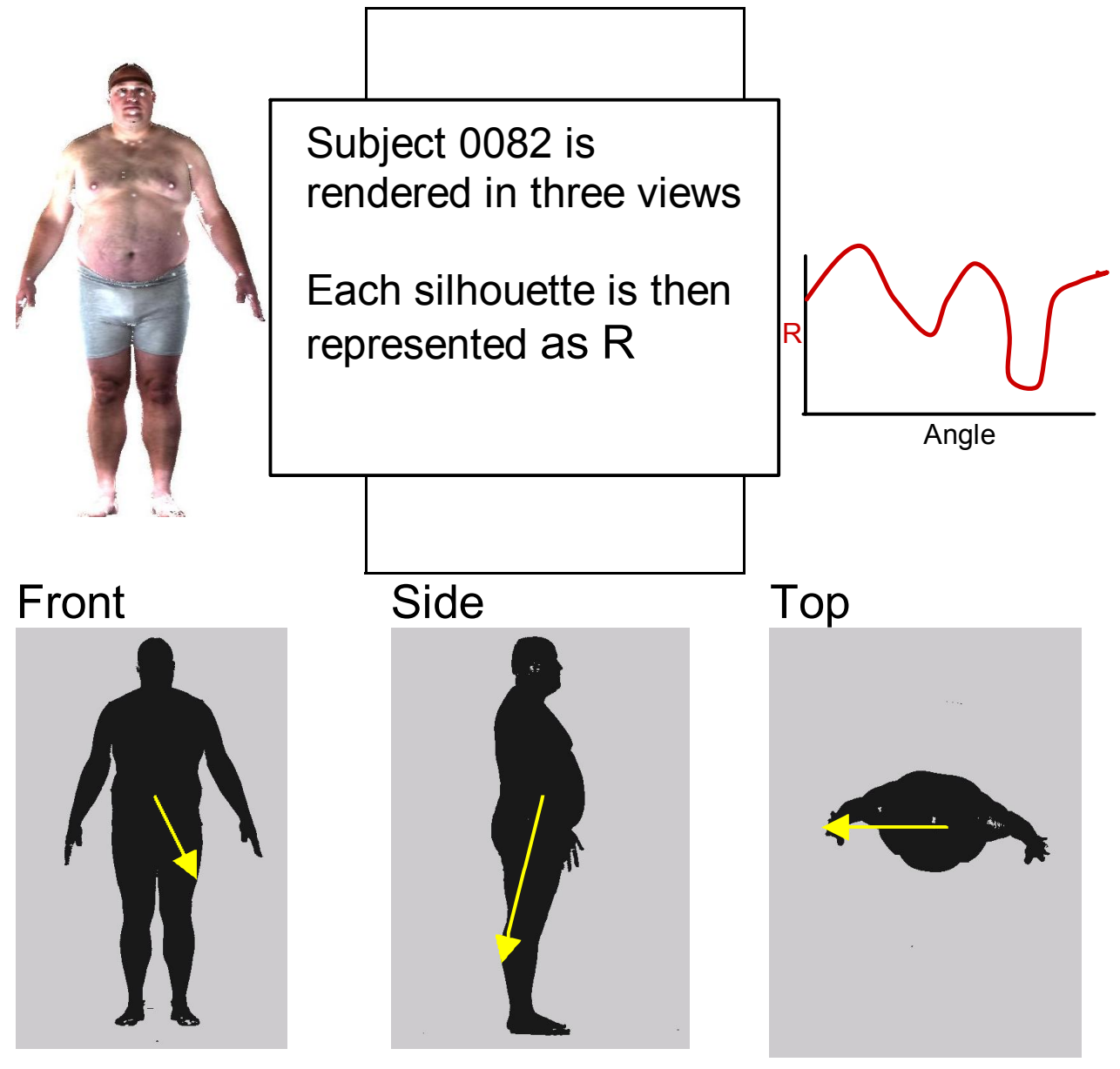

Figure 4. Subject 00082 is rendered in three silhouette views

## HEAD SHAPE DESCRIPTOR

We now describe two methods for creating descriptors based on human head shape:

### PCA Based

In this method we neglected the effect of facial expression. By cutting part of the facial grid from the whole CAESAR body grid using the landmark points 5 and 10 as shown in Figure 5 and listed in Table 1. Table 1 list all the numbers and names of landmark points used in our 3D face recognition study. The new generated facial grid for some of the subjects with two different views is shown in Figure 6. In the case of people standing the minimum number of grid points is 2445 and the mean number is 5729. For the case of people sitting the minimum number of grid points in the facial surface is 660 and the mean number is 4533. This shows that the grid is very coarse for some of the subjects in the seated pose.

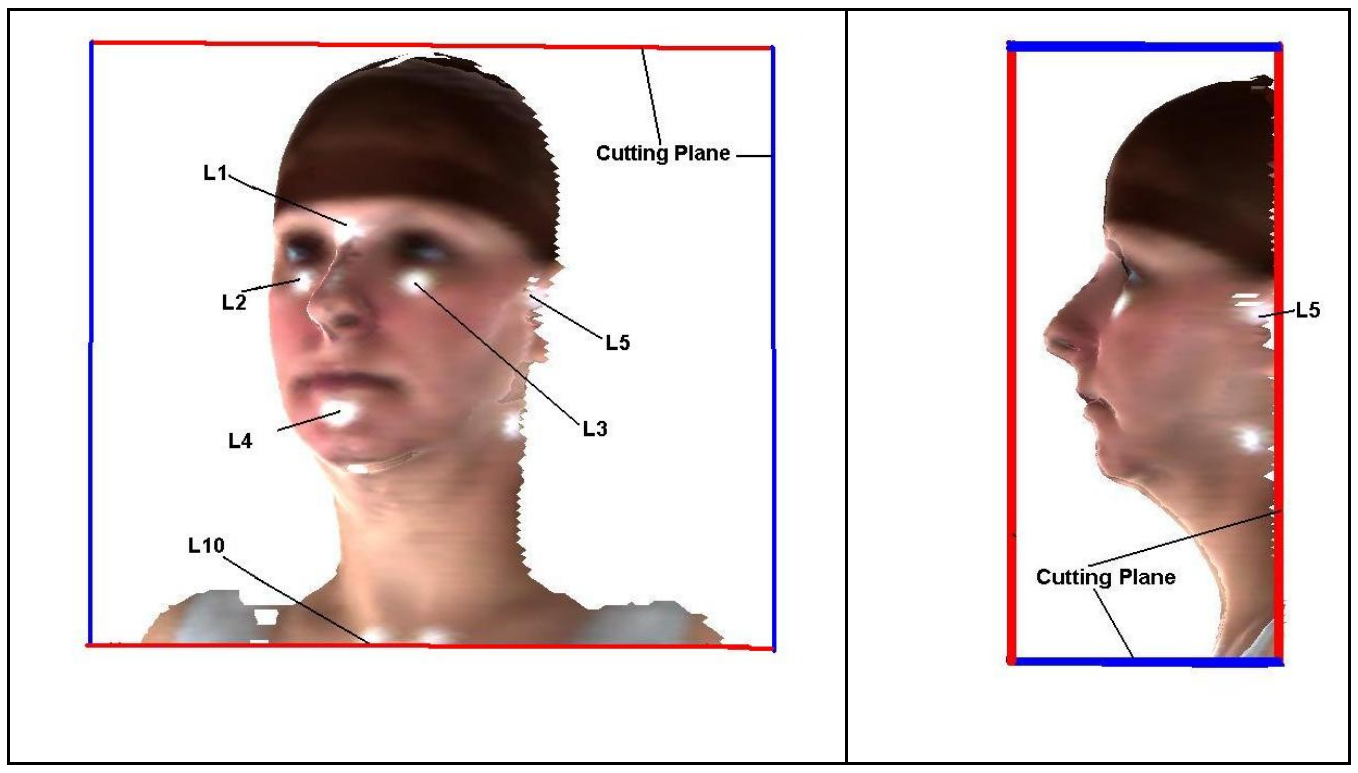

Figure 5. Landmark points 1, 2, 3, 4, 5 and 10. Vertical and horizontal lines are the cutting plane

Table 1. Numbers and names of landmark points used in our 3D face

| | | | |
|---|---|---|---|
| 1 | Sellion | 2 | Rt Infraobitale |
| 3 | Lt Infraobitale | 4 | Supramenton |
| 5 | Rt.Tragion | 6 | Rt. Gonion |
| 7 | Lt. Tragion | 8 | Lt. Gonion |
| 10 | Rt. Clavicale | 12 | Lt.Clavicale |

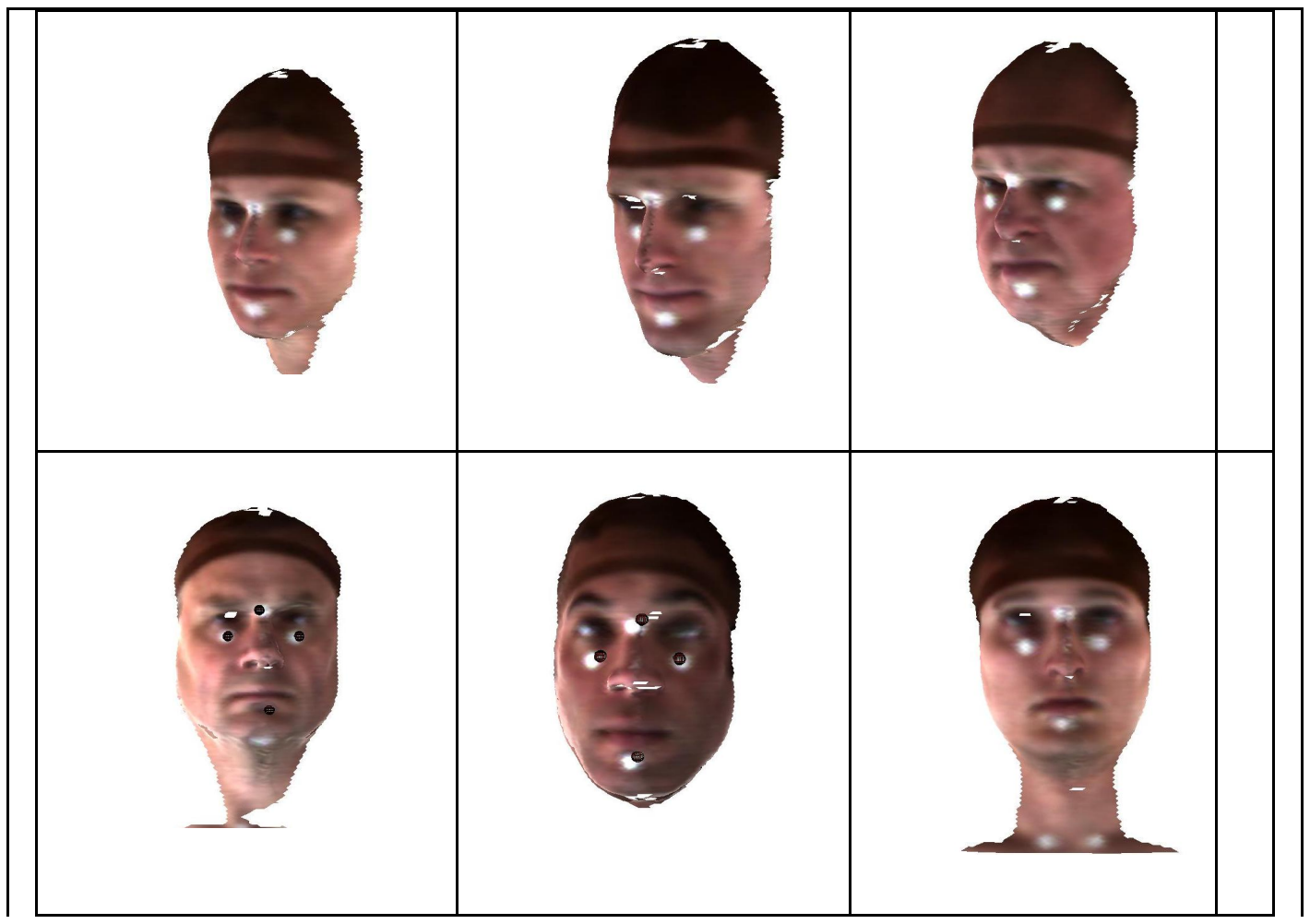

Figure 6. Facial surfaces after the cut from the CAESAR body in two different views.

Next, we use four anthropometric landmark points (L1, L2, L3, L4) as shown in Figure 5, located on the facial surface, to properly position and align the face surface using an iterative method. There is some error in alignment and position because of error in measurements of the position of these landmark points. This Max error was 15 mm, obtained by taking the difference of distance between landmark points |L1- L2| and between |L3 –L4| for subjects standing compared to subjects sitting. Then we interpolate the facial surface information and color map on a regular rectangular grid whose size is proportional to the distance between the landmark points L2 and L3 ( d=| L3 - L2 | ) and whose grid size is 128 in both directions. We use a cubic interpolation and handle missing values with the nearest neighbor method when there are voids in the original facial grid. For some of the subjects there are large voids in the facial surface grids. Figure 7 shows the facial surface and the new rectangular grid.

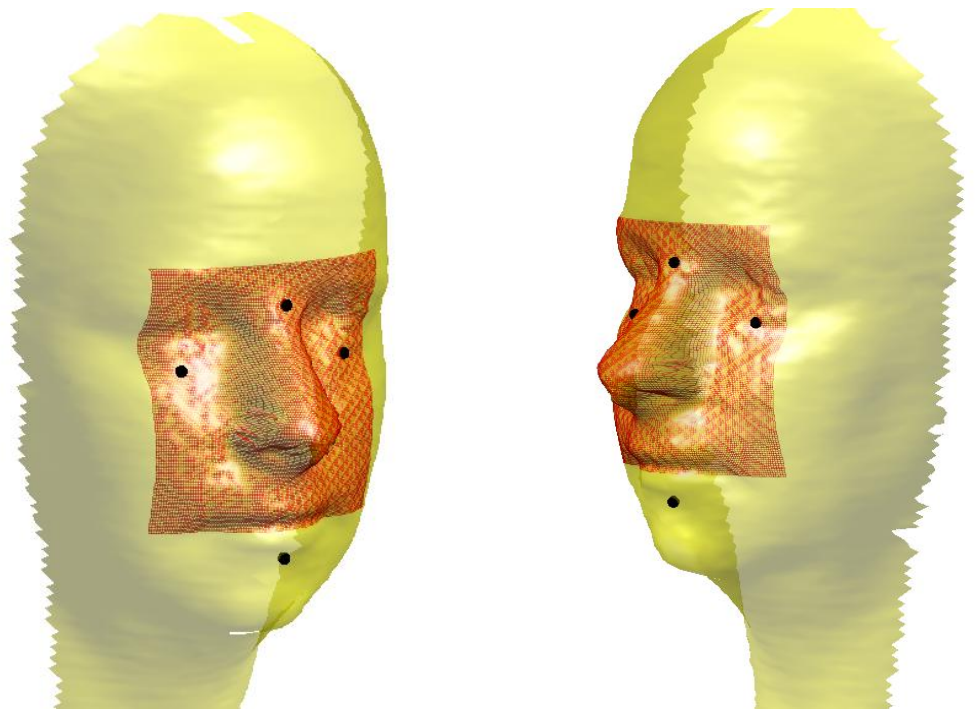

Figure 7. Shows the new facial rectangular grid for two subjects

We properly positioned and aligned the facial surface and then interpolated the surface information on a regular rectangular grid whose size is proportional to the distance between the landmark points. Next we perform Principal Component Analysis (PCA) on the 3D surface and similarity based descriptors are created. In this method the head descriptor is only based on the facial region. The PCA recognition method is a nearest neighbor classifier operating in the PCA subspace. The similarity measure in our study is based on L1 distance and Mahalanobis distance.

To test how well the PCA based descriptor performs, we studied the identification between 200 standing and sitting subjects. The CMC at rank 1 for the study is 85%. More details about this descriptor are described in the paper by [Godil 2004]

## Spherical Harmonics Based

In the second method the 3D triangular grid of the head is transformed to a spherical coordinate system by a least square approach and expanded in a spherical harmonic basis as shown in Figure 8. Since the CAESAR head grid has large voids in the top of the head and also because of cutting the grid at the neck there is circular hole. Since these holes are not filled properly, we have a convergence problem with 10% of the head grids. The main advantage of the Spherical Harmonics Based head descriptor is that it is orientation and position independent. In the near future we plan to fix this problem using a method which fills voids. The spherical harmonics based descriptor is then used with the L1 and L2 norm to create similarity measure.

To test how well the Spherical Harmonics Based head descriptor performs, we studied the identification of the human head between 220 standing and sitting subjects. The CMC at rank 1 for the study is 94%.

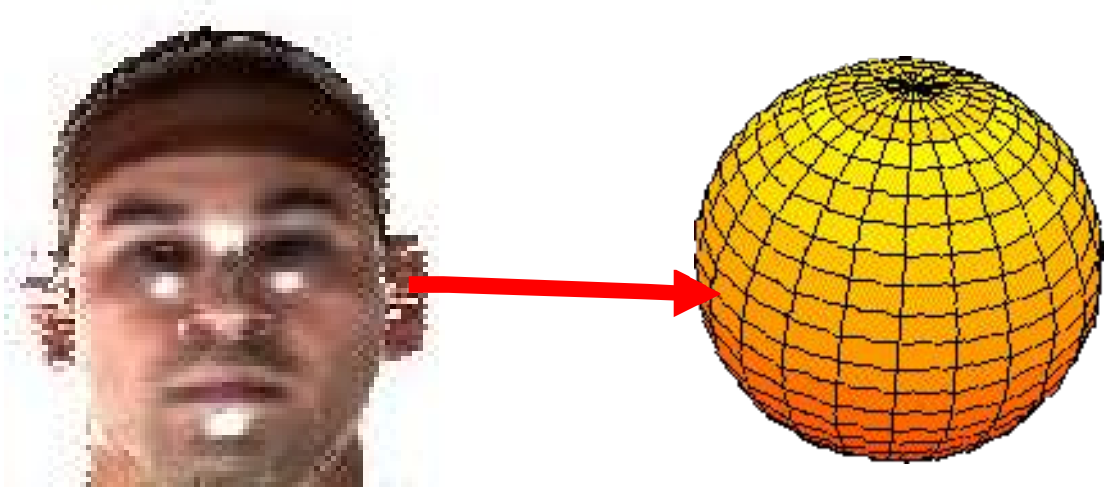

Figure 8. 3D head grid is mapped into a sphere

# RESULTS

## Retrieval Results

The web based interface enables us to select a particular body, or a random body or bodies, based on some criteria such as weight, age, height, etc as shown Figure 9. Subsequently, we can perform similarity based retrieval based on a single descriptor (out of the four descriptors). Using four descriptors allows users to select the best descriptor for their application, such as the use of head descriptor for helmet or eyeglasses design. The partial results from a body shape based similarity retrieval for subject number 16270 are shown Figure 10.

The partial results from a head shape PCA based similarity retrieval for subject number 00068 are shown Figure 11 and for subject number 00014 are shown in Figure 12.

The initial results show that the results and amount of time for retrieval are very reasonable.

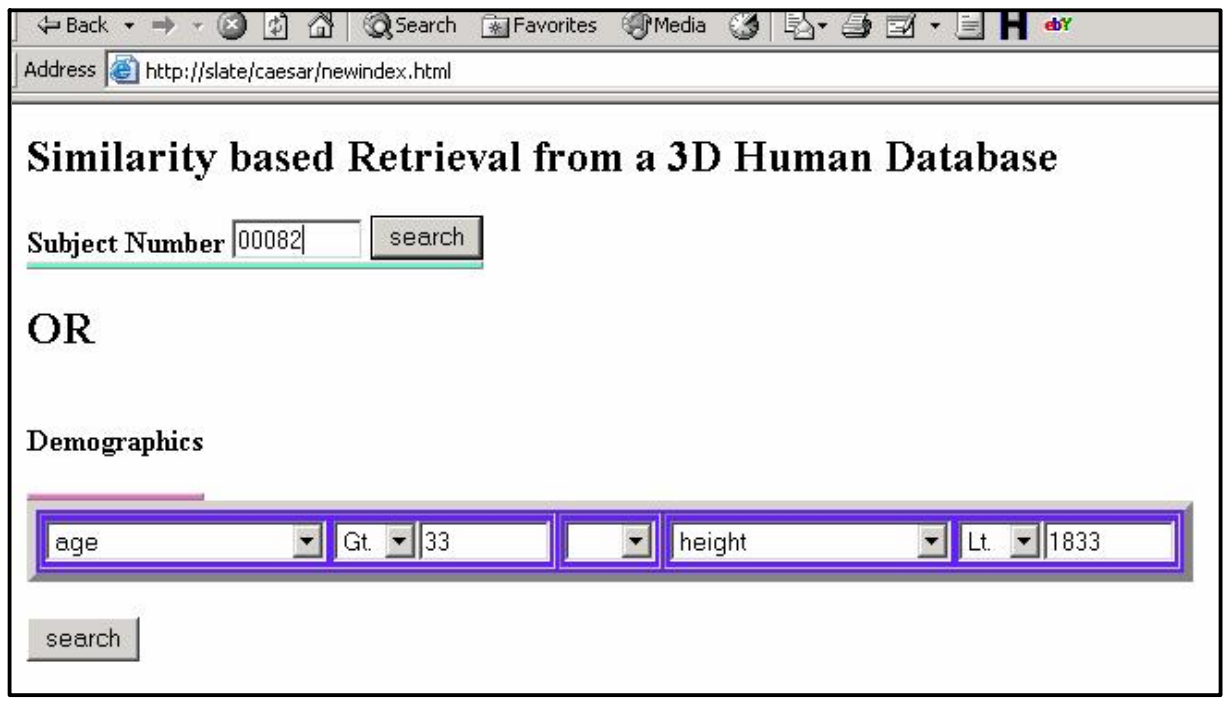

Figure 9. The web based interface allows one to select a particular body, or a random body

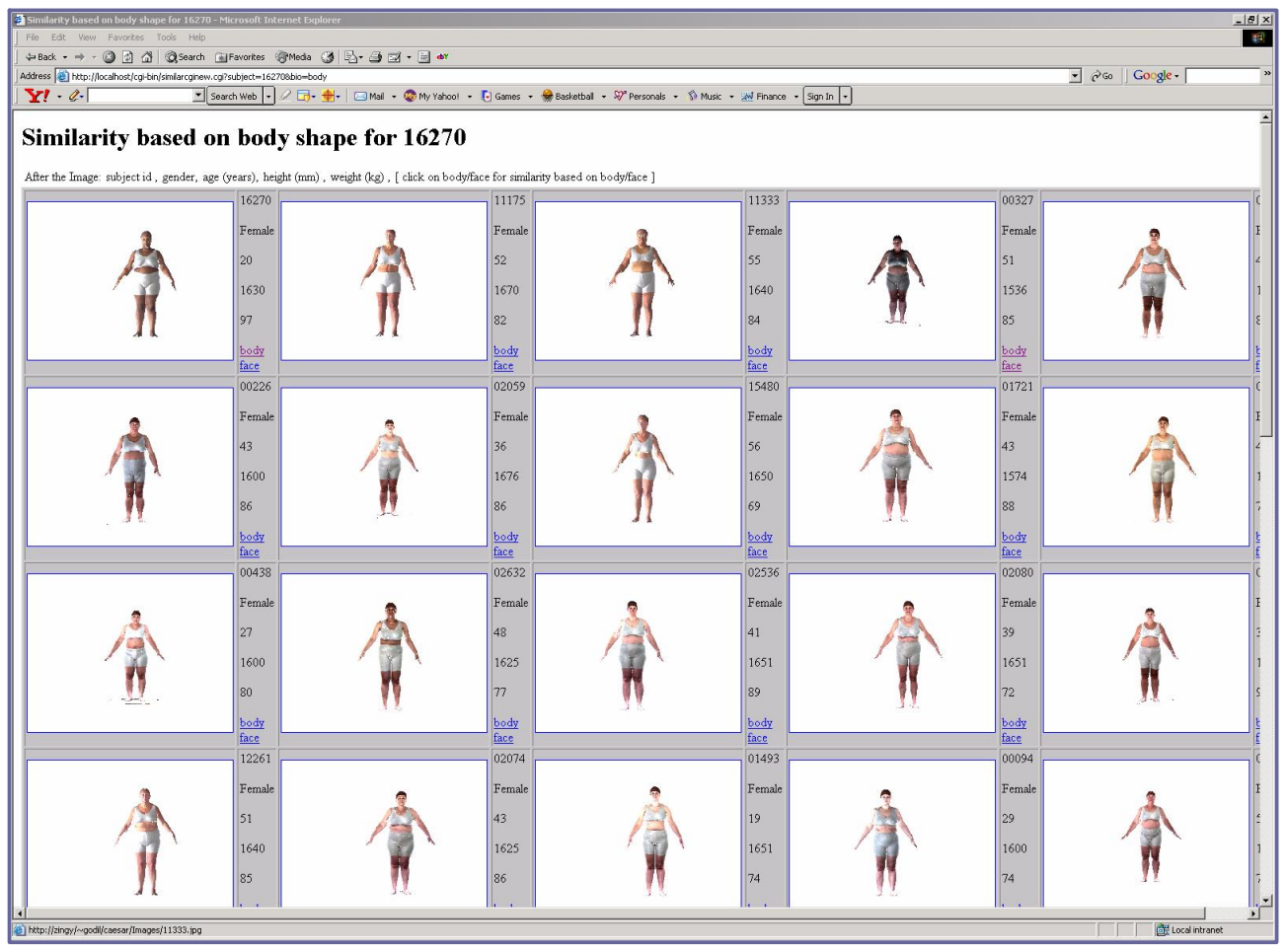

Figure 10. Similarity based retrieval for 16270 based on body shape

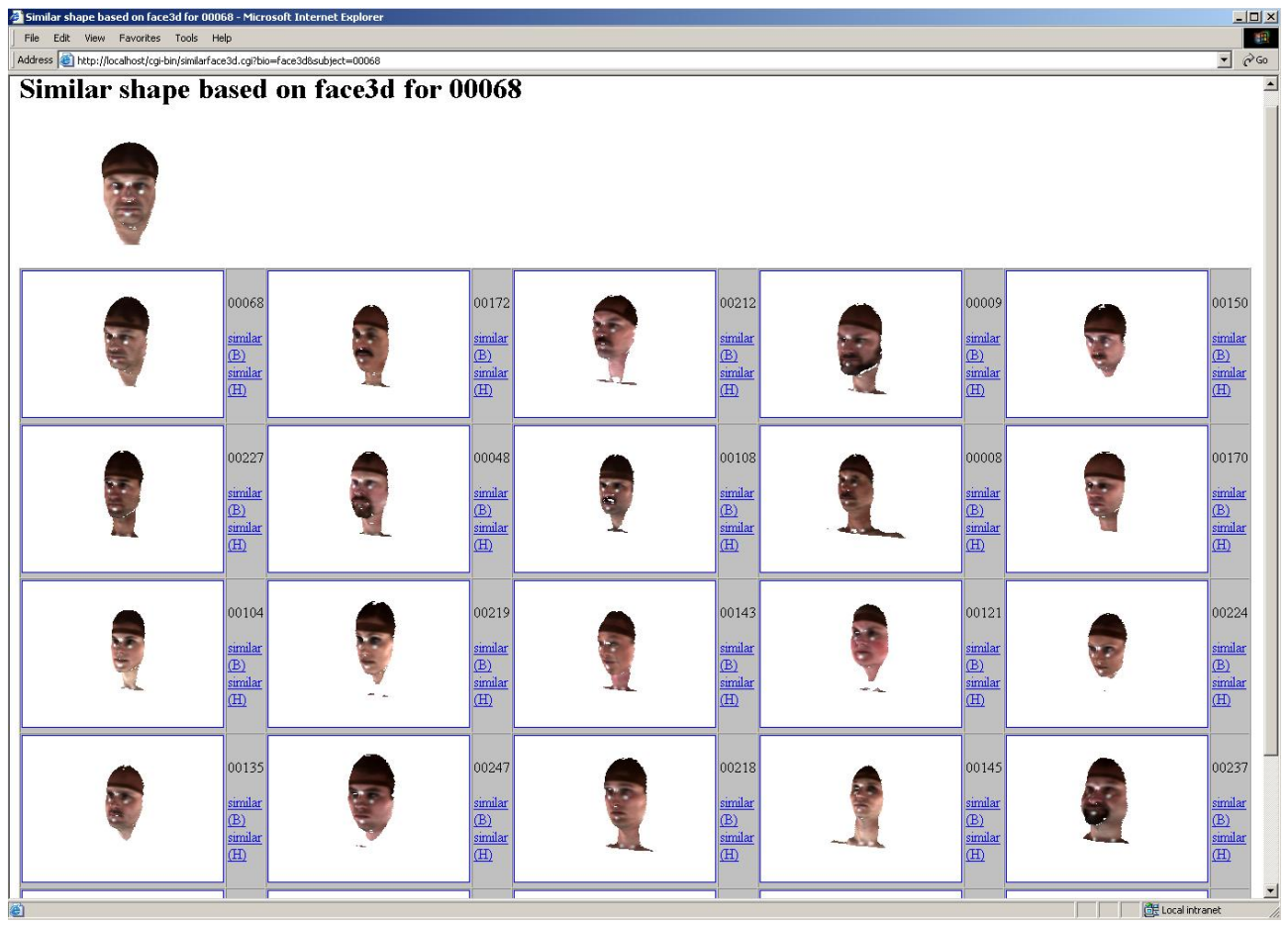

Figure 11. Similarity based retrieval for 00068 based on PCA facial shape

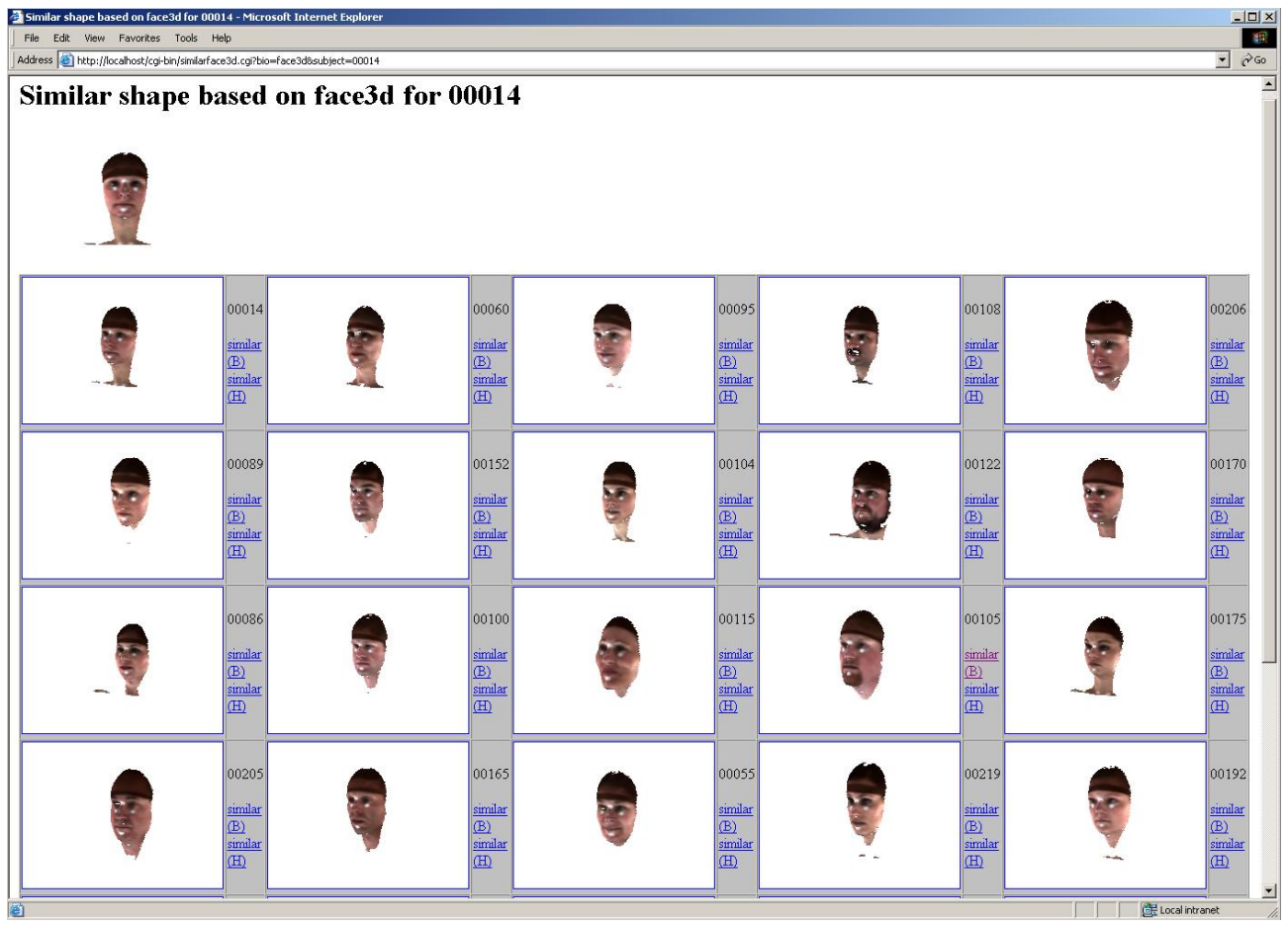

Figure 12. Similarity based retrieval for 00014 based on PCA facial shape

## Clustering Results

We have used the compact body and head descriptors for clustering. Clustering is the process of organizing a set of bodies/heads into groups in such a way that the bodies/heads within the group are more similar to each other than they are to other bodies belonging to different clusters. Many methods for clustering are found in various communities; we have tried a hierarchical clustering method. We then use Dendrogram which is a visual representation of hierarchical data to show the clusters. The Dendrogram tree starts at the root, which is at the top for a vertical tree (the nodes represent clusters). Figure 12 shows the Agglomerative Clustering of Body Shape Distances descriptor with number of clusters = 100 and Figure 13 shows the same with number of clusters = 30.

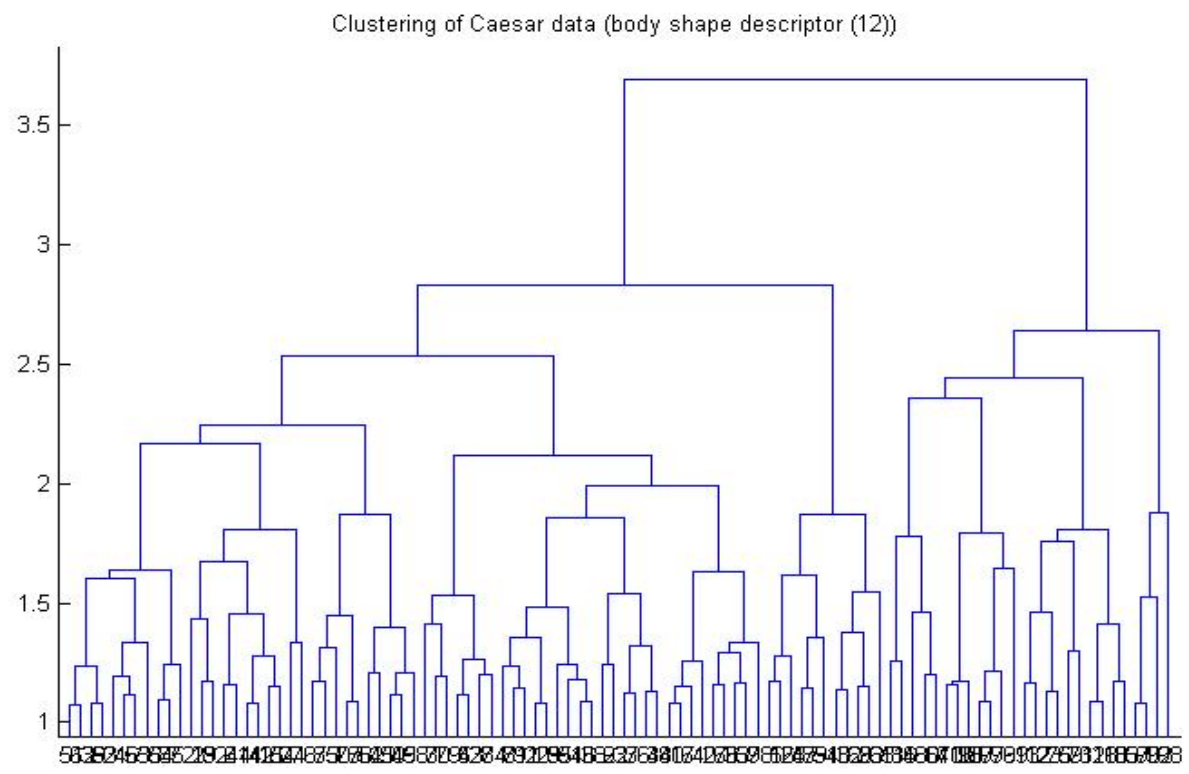

Figure 12. Agglomerative Clustering of Body Shape Distances descriptor (number of clusters=100 )

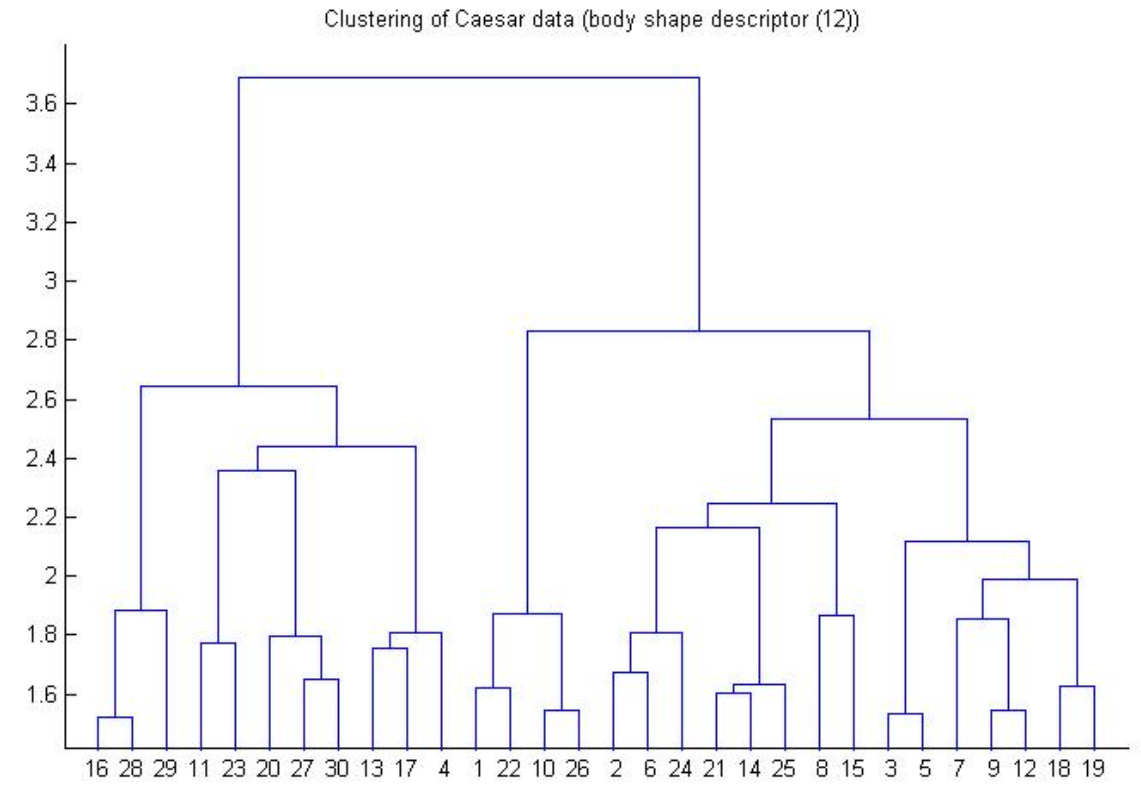

Figure 13. Clustering of Body Shape Distances descriptor (number of clusters=30 )

## Conclusion

We have developed four methods for searching the human database using similarity of human body and head shape. Based on some of our initial observations and from our CMC results for an identification study between 200 standing subjects and 200 sitting subjects, it can be said that the body and head descriptors represent the CAESAR bodies quite accurately. We have seen that our approach performs the similarity based retrieval and clustering in a reasonable amount of time and therefore, has potential to be a practical approach.

## ACKNOWLEDGMENTS

We would like to thank Dr. Kathleen Robinette of Wright - Patterson Air Force Base, Dayton, USA for providing us the CAESAR Anthropometry database.

## CONTACT

Afzal Godil can be contacted at afzal.godil@nist.gov

NIST, 100 Bureau Dr, MS 8940, Gaithersburg, MD 20899